\documentclass[twoside,twocolumn,10pt]{article}
\usepackage{wscg}           % includes a number of other packages (e.g., myalgorithm)
\RequirePackage{ifpdf}
\ifpdf
 \RequirePackage[pdftex]{graphicx}
 \RequirePackage[pdftex]{color}
\else
 \RequirePackage[dvips,draft]{graphicx}
 \RequirePackage[dvips]{color}
\fi
\RequirePackage{amsmath}
\usepackage{nopageno}  
\usepackage[switch]{lineno}
\usepackage{upgreek} 		% fuer griechische Vektoren, z.B. \mu-Vektor
\usepackage{bm}				% fuer fette griechische Vektoren

\title{Impact of PCA-based preprocessing and different CNN structures on deformable registration of sonograms}

\author{
\parbox{0.25\textwidth}{\centering
Christian Schmidt\\[1mm]
Westf{\"a}lische Hochschule {University of Applied Sciences}\\
Neidenburger Strasse 43\\
45897 Gelsenkirchen\\
Germany\\[1mm]
christian.schmidt\\@w-hs.de
}
\hspace{0.05\textwidth}
\parbox{0.25\textwidth}{\centering
Heinrich Martin Overhoff\\[1mm]
Westf{\"a}lische Hochschule {University of Applied Sciences}\\
Neidenburger Strasse 43\\
45897 Gelsenkirchen\\
Germany\\[1mm]
heinrich-martin.overhoff\\@w-hs.de
}
}

\usepackage{url}
\makeatletter
\def\Uslash{\mathbin{\mathchar`\/}\@ifnextchar{/}{\kern-.15em}{}}
\g@addto@macro\UrlSpecials{\do \/ {\Uslash}}
\def\Ucolon{\mathbin{\mathchar`:}\@ifnextchar{/}{\kern-.1em}{}}
\g@addto@macro\UrlSpecials{\do : {\Ucolon}}
\makeatother

\begin{document}

\twocolumn[{\csname @twocolumnfalse\endcsname

\maketitle  % full width title

\begin{abstract}
\noindent
Central venous catheters (CVC) are commonly inserted into the large veins of the neck, e.g. the internal jugular vein (IJV). CVC insertion may cause serious complications like misplacement into an artery or perforation of cervical vessels. Placing a CVC under sonographic guidance is an appropriate method to reduce such adverse events, if anatomical landmarks like venous and arterial vessels can be detected reliably. This task shall be solved by registration of patient individual images vs. an anatomically labelled reference image. 
In this work, a linear, affine transformation is performed on cervical sonograms, followed by a non-linear transformation to achieve a more precise registration. Voxelmorph (VM), a learning-based library for deformable image registration using a convolutional neural network (CNN) with U-Net structure was used for non-linear transformation. The impact of principal component analysis (PCA)-based pre-denoising of patient individual images, as well as the impact of modified net structures with differing complexities on registration results were examined visually and quantitatively, the latter using metrics for deformation and image similarity. Using the PCA-approximated cervical sonograms resulted in decreased mean deformation lengths between 18\% and 66\% compared to their original image counterparts, depending on net structure. In addition, reducing the number of convolutional layers led to improved image similarity with PCA images, while worsening in original images. Despite a large reduction of network parameters, no overall decrease in registration quality was observed, leading to the conclusion that the original net structure is oversized for the task at hand.

%Overall, satisfying registrations of the IJV were achieved with all net structures. Although little impact of net structures on registration performance was observed, their varying deformation properties can be used to further enhance registrations.

%When applying a deformable transfomation to already affinely pre-registered images of regular structure, it can be assumed that only a small amount of deformation is needed to align those images. Using deep neural networks for this registration, overfitting can lead to unnecessarily large deformations because of factors like image noise and artifacts. In this work, principal component analysis (PCA), a statistical method which reduces the dimensionality of data, was used to approximate a sonogram data set, by linear combination of its principal components, which smoothens the images and reduces noise. Using the PCA-approximated neck sonograms resulted in decreased mean deformation lengths between 18\% and 66\% compared to their original image counterparts, depending on net structure. In addition, reducing the number of convolutional layers led to improved image similarity with PCA images while worsening in original images. Despite a large reduction of network parameters, no overall decrease in registration quality was observed, leading to the conclusion that the original net structure is oversized for the task at hand.
\end{abstract}

\subsection*{Keywords}
Medical image registration, deformable registration, sonograms, Voxelmorph, CNN

\vspace*{1.0\baselineskip}
}]

%%%%%%%%%%%%%%%%%%%%%%%%%%%%%%%%%%%%%%%%%%%%%%%%%%%%%%%%%%%%%%%%%%%%%%%%%%%%%

\section{Introduction}

\copyrightspace

\begin{figure*}
	\centering
	\includegraphics[width=0.77\textwidth]{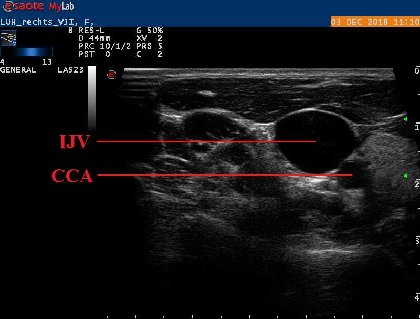}
	\caption{\textbf{Example image of an original cervical sonogram. Internal jugular vein (IJV) and common carotid artery (CCA) are labelled.}}
	\label{anatomy}
\end{figure*}

Placement of a central venous catheter (CVC) is a procedure that carries risk for multiple complications, e.g., arterial puncture of the common carotid artery has an occurrence rate of $6\% - 9\%$~\cite{bee03}. This work aims to further improve the ultrasound guided CVC placement into the internal jugular vein (IJV) (Fig.~\ref{anatomy}), by detecting the IJV and indicating a needle target position in a manually acquired patient individual ultrasound image. The needle target position in such an image is to be determined by automated, computer-based analysis. 
It is assumed that an optimal needle target position is defined in a reference image. The task at hand is to map the optimal needle target position onto patient individual images. In order to realize this mapping, the patient individual images are to be registered vs. the reference image.

Overfitting occurs when a model learns the training data well, but does not generalize the acquired information to new data. This is an issue in medical machine learning applications, since these datasets are usually small compared to the complexity of deep neural network structures, or due to low signal-to-noise ratio in the data. 

The hypothesis of this work is: An improvement of the signal-to-noise ratio in image data and a systematic reduction of network size can yield improved registration results with overall less deformation, and thus a more regular registration field. 
In this work, principal component analysis (PCA) is used for noise reduction, and Voxelmorph, a U-Net-based convolutional neural network (CNN), is used as a reference network structure. 
To evaluate the hypothesis, three models for both image types are parameterized for a fine registration of affinely pre-registered ultrasound images of the human neck. 
The main tasks of this work are:
\begin{itemize}
%	\item only few clinical images are available; these images contain different anatomical structures with varying image contrast. Therefore, the original ultrasound image is reduced in size to a region of interest (ROI) that contains mainly the VJI. This is done by feature based image segmentation and a following affine pre-registration of the images.
	
\item reduce the size of original ultrasound images to a region of interest (ROI) that contains mainly the IJV. This shall be done by feature-based image segmentation. Subsequently, apply affine pre-registration to the ROI images. Because only few clinical images are available, and those have varying anatomical structures and image contrast, this procedure shall make the
defomable image registration less error-prone.
	\item Perform a PCA on the image data set and approximate it by linear combination of the most relevant principal components.
	\item Train neural networks with three different structures and different number of free parameters to register image pairs (non-linear, deformable transformation). For each net structure, train two versions, one for original images, and one for PCA-approximated images.
	\item Quantitatively analyze the impact of the number of net parameters and the image type on the registration result, using evaluation metrics for deformation and image similarity.
\end{itemize} 
\begin{figure*}
	\centering
	\includegraphics[width=\textwidth]{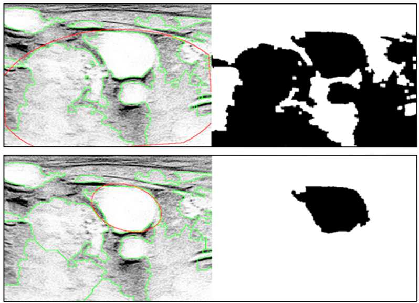}
	\caption{\textbf{$n$ largest objects after binarization are delineated with green contours, the object identified as correct is marked with a red ellipse (left). Top row shows objects before, bottom row objects after watershed-transformation. Corresponding binary images after thresholding with  ${g}_\mathrm{thresh}$ are shown on the right. Sonograms are dispayed with inverted grayscale pixel values for better visibility.}}
	\label{segmentation}
\end{figure*}
\begin{figure*}
	\centering
	\includegraphics[width=\textwidth]{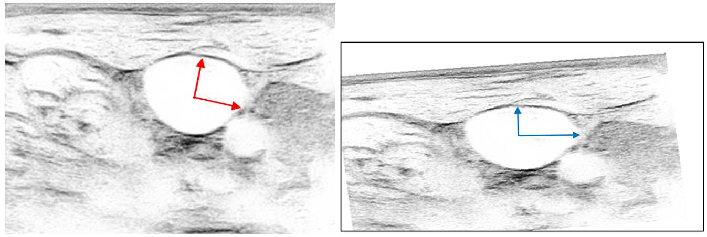}
	\caption{\textbf{Example sonogram of the IJV before (left) and after (right) affine transformation. Ellipse parameters resulting from the preceding feature-based segmentation, are used to translate ($\mathbf{x}_{OR}$), rotate ($-\varphi$) and scale ($s_x$, $s_y$) the images to coarsely pre-register them for the subsequent deformable deformation by the CNN.}}
	\label{affinetrans}
\end{figure*}

\section{Related Work}
Many different methods for deep learning-based medical image registration have been proposed in the past. For rigid transformations in particular, deep reinforcement learning (RL) techniques~\cite{mni15} have gained some popularity.~\cite{lia17} proposed a RL strategy for rigid 3D-3D registrations in computer tomography images, which is based on finding the optimal sequence of motion actions (rotations and translations) for image alignment. Since RL networks are constrained to low dimensionality of outputs, they have been used almost exclusively for rigid registrations, since those can be expressed by a small number of transformation parameters. With the rise of networks, which can directly estimate deformation vector fields (DVF) and are not constrained to rigid transformations, RL-based methods fell out of favor in recent years~\cite{fu20}. 

\begin{figure*}
	\centering
	\includegraphics[width=\textwidth]{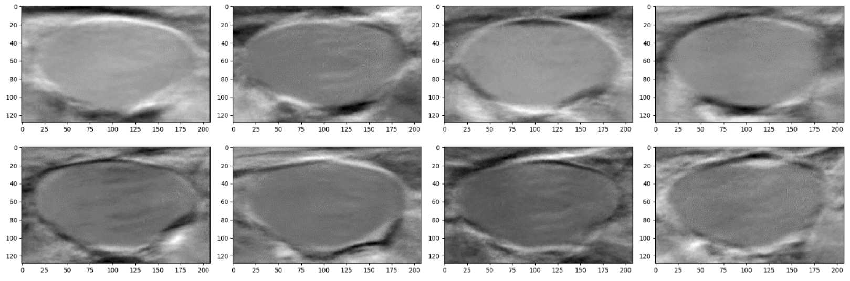}
	\caption{\textbf{First $q = 1 \ldots 8$ principal component images $\mathbf{G}(\mathbf{y}_1)$ through $\mathbf{G}(\mathbf{y}_8)$  from performing a PCA on the ultrasound image data set (top row $\mathbf{G}(\mathbf{y}_1) \ldots \mathbf{G}(\mathbf{y}_4)$ , bottom row $\mathbf{G}(\mathbf{y}_5) \ldots \mathbf{G}(\mathbf{y}_8)$).}}
	\label{component}
\end{figure*}
Networks, which directly estimate the DVF can be classified into supervised and unsupervised methods. Supervised networks require ground truth transformations (either DVF, in case of deformable registration, or rigid transformation parameters). Ground truth transformations can be obtained by artificially de-aligning images with random rotations and translations~\cite{sal19, epp18} or by using traditional, non-learning methods to register image pairs and use the resulting DVFs as ground truth for training~\cite{sen18}.
\begin{table*}
	
	\begin{center}
		\begin{tabular}{c|c| c| c| c |c| c| c| c| c| c} 
			%\hline
			$q$ & 1 & 3 & 5 & 7 & 9 & 11 & 13 & 15 & 17 & 19 \\ 
			\hline
			$cEVR$ & 0.12 & 0.31 & 0.46 & 0.55 & 0.62 & 0.67 & 0.71 & 0.75 & 0.77 & 0.79 \\ 
			%\hline
		\end{tabular}
		\caption{\textbf{Cumulative explained variance ratio $cEVR$ by number of first $q$ principal components.}}
		\label{table}
	\end{center}
\end{table*}
Lack of  medical datasets with known ground truth DVFs led to a rising demand for unsupervised networks. With the introduction of spatial transformer networks~\cite{jad15}, calculating image similarity losses was made possible during training. This is achieved by warping the estimated DVF with the input image and comparing the resulting image with the reference image. These networks do not require supervision by ground truth annotations and in addition to image similarity loss, employ a regularization loss term, to ensure smooth and anatomically plausible transformations~\cite{zha18, bal19}.

Looking at image modalities, DL-based registration of ultrasound (US)-images only played a subordinate role in recent research, despite the high prevalence of sonography in clinical practice. A review of current publications in medical image registration~\cite{bov20} found, that US-images were only used in about $5\%$ of research papers with the topic of DL-based medical image registration, while magnetic resonance imaging (MRI) ($52\%$) and computer tomography (CT) ($19\%$) dominated the field. This is mainly due to higher availability of public MR training datasets and the fact, that the majority of papers examined registrations of brain images, which are predominantly recorded in MR and CT scans. US-images were most often used in multi-modal registration tasks~\cite{hu18, yan18}, in which, e.g., pre-procedural MR scans were aligned with intra-procedural US-images.

\section{Proposed solution}
\subsection*{Segmentation and affine pre-registration}
Firstly, the original ultrasound image is being cropped and the interface of the ultrasound machine is removed. To segment the image into foreground and background, it is binarized using a binarization threshold value  ${g}_\mathrm{thresh}$.
Subsequently, everything but the $n$ largest objects are removed, to filter out structures that are too small to reasonably be considered. A watershed transform is implemented (Fig.~\ref{segmentation}), since as  ${g}_\mathrm{thresh}$ increases, the increasingly forming clusters need to be separated, in order to be detected as individual entities. Choosing the correct IJV-object out of the remaining $n$, is done via the $y$-coordinate of the object's center and the distance between the common carotid artery (CCA) and the respective object. The object center's $y$-coordinates can be utilized because the distal location of the IJV is roughly the same for all subjects. Additionally, since CCA and IJV are in close anatomical proximity, all objects outside of a certain distance to the CCA can be excluded.

To obtain a smoother contour and an object which is geometrically parameterizable, the correctly identified object is approximated with an ellipse. The ellipse parameters major and minor axis length ($a$, $b$), and major axis rotation angle  vs. the $x$-axis ($\varphi$) are extracted from the ellipse approximation. These parameters are subsequently used in the affine transformation (pre-registration).

This affine transformation between object IJV ($O$) and reference IJV ($R$) (Fig.~\ref{affinetrans}) was performed as follows: At first, the image is translated by $\mathbf{x}_{OR}$, which aligns the center of the approximated ellipse with the image center. This is also the origin of the new reference coordinate system. Secondly, a rotation of $-\varphi$ degrees is applied to the image. This alignment of the center and rotation angle of the object IJV and the new reference coordinate system can be described as a 2D rigid body transform. Subsequently, image coordinate axes are scaled using the scaling factors $s_x $ and $s_y$. Values for $s_x = \frac{a_R}{a_O}$ and $s_y = \frac{b_R}{b_O}$ are used to match the major and minor axes lengths of object vs. reference IJV ellipses. Finally, a rectangle of size 208 $\times$ 128 pixels around the object center is cropped, to only leave relevant parts of the image for later use as training data.

\subsection*{Principal Component Analysis}
PCA is used to reduce dimensionality in the ultrasound image data set. For this purpose, $p = 81$ pre-registered ultrasound images of the human neck (transversal plane) from 14 different subjects (five to six images per subject) of size $208 \times 128$ pixels are investigated.

PCA is a statistical method, which is used for projecting a $p$-dimensional data set into a $q$-dimensional sub-set $(q~<~p)$, while preserving characteristic data variability~\cite{jol16, kon17}. 
A data set consists of $p$ variables $x_i$, $1\leq i \leq p$, with $n$ observations. Each variable $x_i$ has a mean $\mu_i$ and a variance $\sigma^2_i$ calculated over its $n$ observations. The sum over the variance of all variables is the total variance 
\begin{linenomath}\begin{equation*}
	\sigma^2_\mathrm{total}=\sum_{i=1}^{p}\sigma^2_i
\end{equation*}\end{linenomath}
\begin{figure*}
	\centering
	\includegraphics[width=\textwidth]{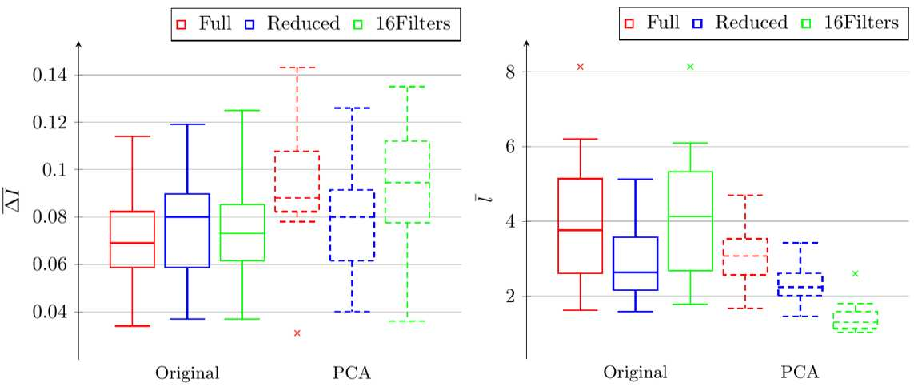}
	\caption{\textbf{Results of registrations with different net structures, using original and PCA-approximated images. Mean of differences of absolute grayscale intensities $\overline{\mathrm{\Delta} I}$ is shown on the left, mean deformation vector length $\bar{l}$ on the right.}}
	\label{Boxplot}
\end{figure*}
Observations $x_{mi}$ of variable $x_i$, $1\leq m \leq n$, are noted as a $n \times 1$ vector $\mathbf{x}_i$. With the observed mean for vector $\mathbf{x}_i$ being $\bm{\upmu}$, the PCA is calculated over centered observations $\mathbf{X}_i=\mathbf{x}_i-\bm{\upmu}$. The eigenvalues $\lambda_j$  of the co-variance matrix are indexed in descending order, they represent variances and fulfill  
\begin{linenomath}\begin{equation*}
	\sigma^2_\mathrm{total}=\sum_{j=1}^{p}\lambda_j
\end{equation*}\end{linenomath}
The PCA yields new variables $\mathbf{y}_j$, the so-called principal components (PCs). PCs are linear combinations of centered observations and have an identical co-variance matrix, i.e., the eigenvalues $\lambda_j$ are the variances of variables $\mathbf{y}_j$. The total variance $\sigma^2_\mathrm{total}$ is identical for PCs, original data, and centered observations, but is distributed differently among the variables. Goal of the PCA is to explain a major part of the total variance with a small number of variables $q$, the cumulative explained variance ratio ($cEVR$) is given by

\begin{linenomath}\begin{equation*}
	cEVR=\frac{\sum\limits_{j=1}^{q}\lambda_j}{\sigma^2_\mathrm{total}}
\end{equation*}\end{linenomath}
The first $q$ of all $p$ new variables $\mathbf{y}_1 \ldots \mathbf{y}_q$ determine the data sub-set, such that
\begin{linenomath}\begin{equation*}
	\mathbf{x}_i \approx \tilde{\mathbf{x}}_i=\sum\limits_{j=1}^{q}\beta_{ij} \mathbf{y}_j + \bm{\upmu}
\end{equation*}\end{linenomath}

To perform the PCA, pre-registered ultrasound images of the human neck $\mathbf{G}_i$ are reshaped into column vectors $\mathbf{x} \left(\mathbf{G}_i\right)$ with $n = 26624$ observations each. After dimensionality reduction, the above-described image vectors can be reorganized as images  $\mathbf{G}(\tilde{\mathbf{x}}_i) \approx \mathbf{G}$.
The first $q$ PCs (Fig.~\ref{component}) are used to approximate the original dataset. In the upcoming sections, $q = 8$ is used, which accounts for about 58\% of the data's variance (Table~\ref{table})  while reducing 90\% of dimensionality (from $208 \times 128 \times 81$ to $208 \times 128 \times 8$).

\subsection*{Voxelmorph and variation of net structures}		
Voxelmorph (VM)~\cite{bal19}, a learning-based library for deformable image registration, which uses a U-Net-based ~\cite{ron15} net structure, is used to perform the deformable sonogram registrations. An atlas-based registration approach is used in this work; thus, an image pair consists of a varying patient individual image (moving image \textit{m}) and a reference image (fixed image \textit{f}). 

Voxelmorph uses a two-part loss function
\begin{linenomath}\begin{equation*}
J = \mathcal{L}_{\mathrm{sim}}(f, m\circ\phi)+\gamma \mathcal{L}_{\mathrm{smooth}}(\phi),
\end{equation*}\end{linenomath}
which consists of a similarity term $\mathcal{L}_{\mathrm{sim}}(f, m\circ\phi)$ and a deformation term $\mathcal{L}_{\mathrm{smooth}}(\phi)$.	
The loss function $J$ penalizes differences in grayscale values as well as deformations and is minimized by learning optimal convolutional kernels (filters). When registering an image pair, the network yields pixel-wise displacement vectors  
\begin{linenomath}\begin{equation*}
\mathbf{u} =  
\begin{bmatrix} u_x \\ u_y \end{bmatrix} = \begin{bmatrix} x_f \\ y_f \end{bmatrix} -\begin{bmatrix} x_m \\ y_m \end{bmatrix}, l = \| \mathbf{u} \|
\end{equation*}\end{linenomath}
between moving image \textit{m} and fixed image \textit{f}. The registration field
\begin{linenomath}\begin{equation*}
	\phi=Id+\mathbf{u}
\end{equation*}\end{linenomath}
is formed by adding $\mathbf{u}$ to the identity transform.
The resulting registration field $\phi$ generates a moved image $m\circ\phi$, which is similar to \textit{f}. 

By employing a regularization term, Voxelmorph encourages smooth, diffeomorphic deformations, i.e. deformations which are anatomically reasonable.
$\gamma$ serves as the regularization parameter, in this work we used $\gamma=0.001$ . As $\gamma$ increases, deformation becomes more costly and the resulting deformation field, therefore, becomes more regular (smooth) and vice versa. The resulting registration field $\phi$ is applied to the moving image $m$ by a spatial transformer function, to obtain the moved image $m\circ\phi$ ($m$ warped by $\phi$). 
\begin{figure*}
	\centering
	\includegraphics[width=\textwidth]{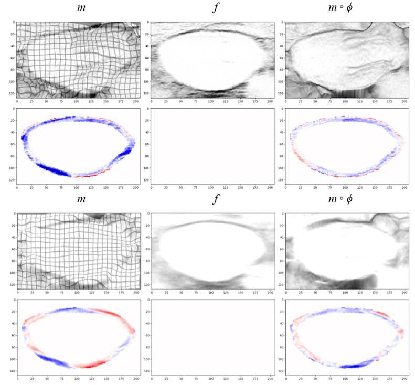}
	\caption{\textbf{Example registration results of original image (top) and PCA-approximation (bottom), using the full net structure. The registration field $\phi$ is warped with a regular square grid and superimposed over the moving image \textit{m}, to show the extent and direction of local deformation which is applied to individual image parts. In addition, values of the pixel-wise pre- and post-registration grayscale difference $\mathrm{\Delta} I$ are displayed color-coded, to illustrate the effect of registration on image similarities (colors ranging from dark red for $\mathrm{\Delta} I = 1$ to dark blue for $\mathrm{\Delta} I = -1$).}}
	\label{PCAcomparison}
\end{figure*}

To examine the effects of reducing the number of free parameters in the CNN by cutting down its size, three net structures are introduced: 
\begin{itemize}
	\item 	The ``full'' structure proposed in the original VM paper, consisting of four encoder and seven decoder convolutional layers with 16 or 32 filters (convolutional kernels) each (16, 32, 32, 32 | 32, 32, 32, 32, 32, 16, 16). This net structure contains about 110,000 parameters.
	\item The ``reduced'' structure. Two encoder and decoder layers are removed for the second configuration (16, 32 | 32, 32, 32, 16 16), resulting in about 53,000 parameters, a reduction of 52\% compared to the full net.
	\item The ``16 filters'' structure contains all eleven convolutional layers, with 16 filters in each layer (16, 16, 16, 16 | 16, 16, 16, 16, 16, 16, 16), resulting in about 33,000 parameters, 70\% less than the full net. 
\end{itemize}
For each net structure, two versions are trained: 
\begin{itemize}
	\item Original image data vs. the reference image, i.e., $\mathbf{G}(\mathbf{x}_i)=\mathbf{G}(\tilde{\mathbf{x}}_i(q=81))$ vs. $\mathbf{G}(\mathbf{x}_\mathrm{(Ref)})=\mathbf{G}(\tilde{\mathbf{x}}_\mathrm{(Ref)}(q=81))$
	\item PCA-approximated image set vs. the reference image, i.e., $\mathbf{G}(\mathbf{x}_i)=\mathbf{G}(\tilde{\mathbf{x}}_i(q=8))$ vs. $\mathbf{G}(\mathbf{x}_\mathrm{(Ref)})=\mathbf{G}(\tilde{\mathbf{x}}_\mathrm{(Ref)}(q=81))$
\end{itemize}
In the upcoming results section however, PCA images are visually evaluated against the PCA-approximation $\mathbf{G}(\tilde{\mathbf{x}}_\mathrm{(Ref)}(q=8))$ of the reference image to account for the difference in brightness and contrast between original and PCA images. 
\begin{figure*}
	\centering
	\includegraphics[width=\textwidth]{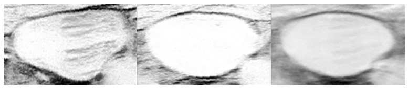}
	\caption{\textbf{Illustration of negative original image features being transferred to PCA-approximations. Reverberation artifacts of original image of subject A (left) appear in the PCA-approximation of subject B (right), even though no such artifacts are present in the original image of subject B (middle).} }	\label{PCAartefakte}
\end{figure*}
\subsection*{Quantitative analysis}
To quantitatively analyze the properties and quality of performed registrations, two evaluation metrics are introduced:
\begin{itemize}
	\item Let $0~\le~I~\le~1$ be the normalized image grayscale intensities, and 
	$\Delta I = I\left( f \right) - I\left( m\circ\phi \right)$ the pixel-wise differences between intensities of fixed image \textit{f} and moved image $m~\circ ~\phi$. Therefore,  $-1~\le~\Delta I~\le~1$ holds. We define  $\overline{\mathrm{\Delta} I}$  as  the mean of absolute grayscale intensity differences $\Delta I$.
	\item Mean deformation vector lengths $\bar{l}$ of the registration field $\phi$, where $l$ measures the pixel-wise deformation (in pixels) that is applied to the moving image \textit{m}. 
\end{itemize}
We use $\overline{\mathrm{\Delta} I}$ and $\bar{l}$ analogously to the similarity term $\mathcal{L}_{\mathrm{sim}}(f, m\circ\phi)$ and the deformation term $\mathcal{L}_{\mathrm{smooth}}(\phi)$ of the Voxelmorph loss function. Since the region around the IJV's contour is of primary importance in this work, $\overline{\mathrm{\Delta} I}$ and $\bar{l}$ are only evaluated in a belt-like along the IJV contour. 

In addition, significances $\alpha$ of metric differences between the above mentioned net and image pair variants are determined with a two-tailed, paired $t$-test.
We used a 70/30 split between training and test data, training the net's parameters on a NVIDIA GeForce RTX 2060 GPU takes 2.5 to 3 minutes, depending on net structure. Registering a single image pair takes 1 to 2 seconds.

\section{Results}

Post-registration mean of absolute intensity differences $\overline{\mathrm{\Delta}I}$ for all net structures and image types are shown in Fig.~\ref{Boxplot}. With original images, an increase in the mean $\overline{\mathrm{\Delta} I}$ of around 12\% can be observed, when using the reduced net instead of the full net (mean $\overline{\mathrm{\Delta} I}: 0.078  \textrm{ vs. } 0.070, \alpha=0.027$). When registering PCA-approximated images however, a 17\% decrease was measured when using the reduced over the full net structure (mean $\overline{\mathrm{\Delta} I}: 0.079 \textrm{ vs. } 0.094, \alpha=0.007$). Comparing the full net structure to their respective 16 filters version showed no significant change in mean $\overline{\mathrm{\Delta}I}$. 

Looking at mean deformation vector lengths $\bar{l}$ (Fig.~\ref{Boxplot}), networks trained with PCA-approximations showed decreases in mean $\bar{l}$ vs. their original image counterpart of 24\% for the full, 18\% for the reduced and 66\% for the 16 filters net structure.
In addition, registrations with PCA-approximated images display the expected smoothing and noise reducing properties, removing unwanted artifacts from the vessel lumen (Fig.~\ref{PCAcomparison}).

\section{Conclusion}
Despite a reduction in net parameters of up to 70\% compared to the originally proposed full net and reducing the mean deformation vector lengths $\bar{l}$ by 18\%~-~66\%, no overall reduction in registration quality was measurable in the downscaled net structures. Specifically, for the combination of reduced net structure with PCA-approximated images, a significant decrease of $\bar{l}$ ($\bar{l} = 2.32$ vs. $2.85, \alpha=0.045$) vs. original images was observed, while $\overline{\mathrm{\Delta}I}$ remained nearly unchanged ($\overline{\mathrm{\Delta}I}= 0.079$ vs. $0.078$). This confirms the hypothesis described in the introduction section, and leads to the conclusion that the full net structure is unnecessarily oversized for the problem at hand. 

The net structure can be reduced in size to diminish problems like overfitting, while also running up to 15\% faster during training compared to the full net structure. In case of images which contain similar, regularly shaped structures, it is recommended to pre-process them with the proposed PCA procedure and employ reduced net structures, to reduce mean deformations and yield more regular registration fields. Since PCA is based on variances, it is highly sensitive to outliers. Thus, noisy images (outliers) in the original data set negatively affect the quality of the principal components, which then in turn affect the approximated PCA images (Fig.~\ref{PCAartefakte}). 

\section{Acknowledgments}
We thank the Federal Ministry of Education and Research (BMBF) Germany, which funded this work in the program ''Gr{\"u}ndungen: Innovative Start-ups f{\"u}r Mensch-Technik-Interaktion'', grant no. 16SV8153.

\end{document}